%% file: paper_main.tex
\documentclass[sigconf]{acmart} 
\AtBeginDocument{%
  \providecommand\BibTeX{{%
    Bib\TeX}}}

\usepackage{amsmath,amsfonts}
\usepackage{algorithmic}
\usepackage{algorithm}
\usepackage[caption=false,font=normalsize,labelfont=sf,textfont=sf]{subfig}
\usepackage{textcomp}
\usepackage{stfloats}
\usepackage{url}
\usepackage{balance}
\usepackage{verbatim}
\usepackage{mathtools}
\usepackage{multirow}
\usepackage{multicol}
\usepackage{glossaries}
\usepackage{graphicx}
\usepackage{array}
\usepackage{colortbl}
\usepackage{xcolor}
\usepackage{arydshln}
\usepackage{adjustbox}
\usepackage{cleveref}
\usepackage{pifont}
\usepackage{balance}

\newcommand{\cmark}{\ding{51}}%
\newcommand{\xmark}{\ding{55}}%

\definecolor{best}{rgb}{1.0, 0.6, 0.6}
\definecolor{second}{rgb}{0.98, 0.78, 0.57}
\definecolor{third}{rgb}{1.0, 1.0, 0.66}
\definecolor{cvprblue}{rgb}{0.21,0.49,0.74}
\def\BibTeX{{\rm B\kern-.05em{\sc i\kern-.025em b}\kern-.08em
    T\kern-.1667em\lower.7ex\hbox{E}\kern-.125emX}}

\acmConference[ACM MM]{ACM multimedia}{Oct 27--31, 2025}{Dublin, Ireland}

\copyrightyear{2025}
\acmYear{2025}
\setcopyright{cc}
\setcctype{by}
\acmConference[MM '25]{Proceedings of the 33rd ACM International Conference on Multimedia}{October 27--31, 2025}{Dublin, Ireland}
\acmBooktitle{Proceedings of the 33rd ACM International Conference on Multimedia (MM '25), October 27--31, 2025, Dublin, Ireland}\acmDOI{10.1145/3746027.3755446}
\acmISBN{979-8-4007-2035-2/2025/10}

\begin{document}

\title{DGNS: Deformable Gaussian Splatting and Dynamic Neural Surface for Monocular Dynamic 3D Reconstruction}



\author{Xuesong Li}
\affiliation{%
  \institution{Agriculture and Food, \\ Commonwealth Scientific and Industrial Research Organisation,}
  \city{Canberra, ACT}
  \country{Australia}
  }
\email{xuesong.li@csiro.au}

\author{Jinguang Tong}
\affiliation{%
  \institution{College of Engineering Computing \& Cybernetics, \\
  The Australian National University,}
  \city{Canberra, ACT}
  \country{Australia}}
\email{jinguang.tong@anu.edu.au}

\author{Jie Hong}
\affiliation{%
  \institution{Faculty of Engineering, \\
  The University of Hong Kong,}
  \city{Hong Kong SAR}
  \country{China}}
\email{jiehong@hku.hk}

\author{Vivien Rolland}
\affiliation{%
  \institution{Agriculture and Food,}
  \institution{Commonwealth Scientific and Industrial Research Organisation,}  
  \city{Canberra, ACT}
  \country{Australia}}
\email{vivien.rolland@csiro.au}

\author{Lars Petersson}
\affiliation{%
  \institution{Data61,}
  \institution{Commonwealth Scientific and Industrial Research Organisation,}  
  \city{Canberra, ACT}
  \country{Australia}}
\email{lars.petersson@data61.csiro.au}






\input{sec/0_abstract}
\maketitle

\input{sec/1_intro}
\input{sec/2_related_work}
\input{sec/2_xpreliminary}
\input{sec/3_methods}
\input{sec/4_experiments}
\input{sec/5_conclusion}

\bibliographystyle{ACM-Reference-Format}
\balance
\bibliography{main}

\end{document}

%% file: sec/0_abstract.tex
\begin{abstract}
Dynamic scene reconstruction from monocular video is essential for real-world applications. 
We introduce DGNS, a hybrid framework integrating \underline{D}eformable \underline{G}aussian Splatting and Dynamic \underline{N}eural \underline{S}urfaces, effectively addressing dynamic novel-view synthesis and 3D geometry reconstruction simultaneously.
During training, depth maps generated by the deformable Gaussian splatting module guide the ray sampling for faster processing and provide depth supervision within the dynamic neural surface module to improve geometry reconstruction. Conversely, the dynamic neural surface directs the distribution of Gaussian primitives around the surface, enhancing rendering quality. In addition, we propose a depth-filtering approach to further refine depth supervision. Extensive experiments conducted on public datasets demonstrate that DGNS achieves state-of-the-art performance in 3D reconstruction, along with competitive results in novel-view synthesis\footnote[1]{\url{https://benzlxs.github.io/dgns_project}; Jie Hong is the corresponding author.}.
\end{abstract}

%% file: sec/1_intro.tex
\section{Introduction}
\label{sec:intro}
\begin{figure}
    \centering   
    \includegraphics[width=0.85\linewidth]{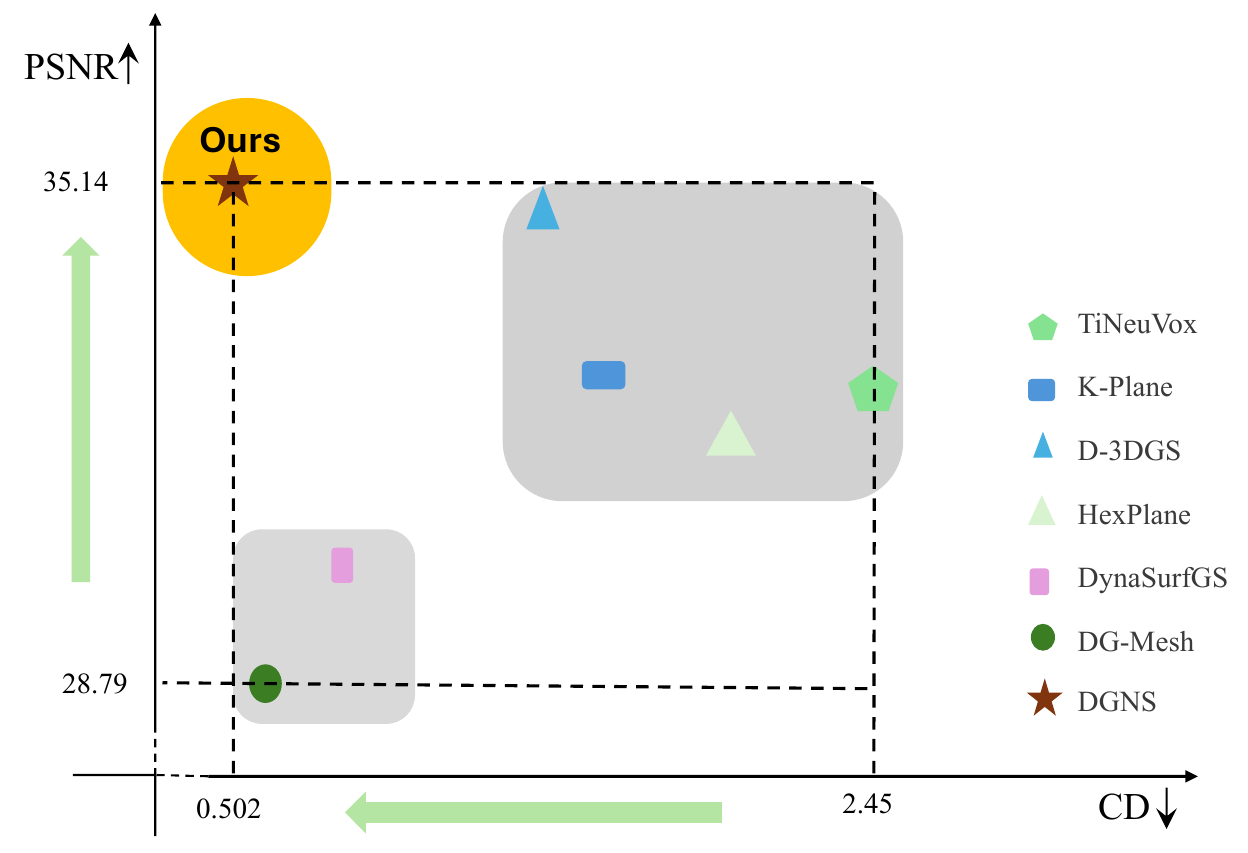}
    \vspace{-1.em}
    \caption{Performance comparison between different methods on the Dg-mesh dataset. The higher PSNR, the better. The smaller CD, the better. Methods representing the best of both should be in the top-left area.}  
    \vspace{-1.5em}
    \label{fig:motivation}       
\end{figure}

Most scenes in our world are dynamic, and achieving dynamic scene reconstruction from monocular video can empower robots or intelligent agents with strong perception capabilities, essential for many real-world applications. Dynamic scene reconstruction primarily involves two key tasks: dynamic novel-view generation and 3D reconstruction. 
In the field of dynamic novel-view synthesis, several approaches~\cite{li2022neural, li2021neural, pumarola2021d, gao2021dynamic} have extended neural radiance fields (NeRF)~\cite{mildenhall2021nerf} by incorporating feature grid planes or implicit deformation fields. Another line of work~\cite{yang2023real, kratimenos2023dynmf, yang2024deformable} models dynamic scenes using explicit Gaussian representations, such as 3D Gaussian Splatting (3DGS)~\cite{kerbl20233d}. While these approaches have achieved promising visual quality, they struggle to accurately recover the 3D geometry of dynamic scenes. For 3D geometry reconstruction in dynamic scenes, some methods~\cite{cai2022neural} have combined deformation fields with implicit surface representations—specifically, signed distance function (SDF) in canonical space. Other methods~\cite{liu2024dynamic, cai2024dynasurfgs} utilize deformable 3DGS to reconstruct dynamic surfaces by introducing strong regularization to ensure Gaussian primitives adhere to the surface. However, these methods encounter difficulties in producing high-fidelity novel-view synthesis. As illustrated in Fig.~\ref{fig:motivation}, current methods tend to excel either in dynamic novel-view synthesis (top-right area) or in 3D reconstruction (bottom-left area), but balancing these two tasks remains an unresolved challenge in dynamic scene reconstruction.

Motivated by the observation that robust geometry guidance can enhance, rather than diminish, rendering quality in static scene modeling~\cite{wang2023adaptive, yu2024gsdf}, we propose a hybrid representation that combines deformable Gaussian and dynamic neural surfaces for dynamic scene reconstruction from monocular video. In this framework, the deformable Gaussian splatting (DGS) module is optimized primarily for appearance reconstruction, while the dynamic neural surfaces (DNS) module focuses on geometry reconstruction. The depth maps generated by the DGS module guide the ray sampling process and provide supervision within the DNS module. At the same time, SDF learned in the DNS module informs the distribution of Gaussian primitives around the surface. To mitigate the noise in depth supervision for the DNS module, we employ Gaussian rasterization to render two types of depth maps—$\alpha$-blended and median depth maps—and introduce a filtering process to create an accurate depth map for supervising depth in the DNS module. As reconstruction from monocular video is a highly under-constrained optimization problem, we introduce normal supervision from the foundation model for both modules. We conducted extensive experiments on two public datasets, where our method outperformed existing approaches on 3D reconstruction with competitive results in novel-view synthesis. Our primary contributions are as follows:
\begin{itemize}
\item We propose a novel hybrid representation combining deformable Gaussian splatting and dynamic neural surfaces, achieving state-of-the-art geometry reconstruction and competitive novel-view synthesis results.
\item We introduce an effective depth-filtering method to enhance depth supervision from Gaussian rasterization.
\item Extensive experiments on public datasets validate the superior performance of our approach over existing methods.
\end{itemize}

%% file: sec/2_related_work.tex
\vspace{-0.5em}
\section{Related work}
\label{sec:related_work}
\subsection{View Synthesis for Dynamic Scene}
View synthesis for dynamic scenes is both challenging and crucial for 3D modeling. NeRF~\cite{mildenhall2021nerf} has shown impressive capabilities in generating high-fidelity novel views for static scenes by using Multi-Layer Perceptrons (MLPs) to model the radiance field, which is then rendered into pixel colors through neural volumetric techniques. Extensions of NeRF to dynamic scenes\cite{li2021neural, pumarola2021d, gao2021dynamic, li2022neural, xiao2025neural, tian2023mononerf, liu2023robust} use time-conditioned latent codes and explicit deformation fields to capture temporal variations. However, the reliance on NeRFs' extensive point sampling along each ray and the computational demands of MLPs limit their scalability for dynamic scenes. To address these limitations, studies have introduced techniques such as hash encoding~\cite{muller2022instant, wang2024masked}, explicit voxel grids~\cite{fridovich2022plenoxels, fang2022fast, xu2023grid, gan2023v4d, guo2024depth}, and feature grid planes~\cite{chen2022tensorf, cao2023hexplane, fridovich2023k, shao2023tensor4d}, which have accelerated training and improved performance in handling dynamic scenes. Another line of research explores geometric primitive rasterization using point clouds~\cite{yifan2019differentiable, aliev2020neural, xu2022point}, offering computational efficiency and flexibility, though these methods often encounter challenges with discontinuities and outliers. More recently, 3DGS~\cite{kerbl20233d} has emerged as a promising approach, leveraging anisotropic 3D Gaussians as rendering primitives. These Gaussians are depth-sorted and alpha-blended onto a 2D plane, enabling high-quality real-time rendering. Various approaches have extended 3DGS to dynamic scenes~\cite{yang2023real, kratimenos2023dynmf, yang2024deformable, katsumata2023efficient,tong2025gs, luiten2023dynamic, guo2024motion}. For example, ~\cite{luiten2023dynamic} introduced dynamic 3DGS by iteratively optimizing the Gaussians per frame, while D3DGS~\cite{yang2024deformable} used deformation fields to model temporal changes in Gaussian distributions. Despite these advancements, such techniques remain primarily focused on novel-view synthesis and often struggle to capture scene geometry accurately, resulting in limitations in high-quality surface extraction.
\vspace{-1.0em}

\subsection{Dynamic Surface Reconstruction}
Reconstructing dynamic surfaces from monocular video is essential for applications such as intelligent robotics and virtual reality. Traditional approaches often depend on predefined object templates~\cite{zuffi20173d, casillas2021isowarp, kairanda2022f} or temporal tracking~\cite{zollhofer2018state, grassal2022neural, feng2021learning}. With advances in neural implicit 3D representations~\cite{park2019deepsdf, mildenhall2021nerf}, methods like LASR~\cite{yang2021lasr} and ViSER~\cite{yang2021viser} reconstruct articulated shapes using differentiable rendering techniques~\cite{liu2019soft}, while BANMo~\cite{yang2022banmo} and PPR~\cite{yang2023ppr} apply NeRF to dynamic scenes, and SDFFlow~\cite{maoneural} models dynamic motion by estimating derivatives of the SDF value. Other approaches utilize RGB-D data to incorporate depth information, improving supervision for dynamic object modeling. Examples include SobolevFusion~\cite{slavcheva2018sobolevfusion}, OcclusionFusion~\cite{lin2022occlusionfusion}, NDR~\cite{cai2022neural}, and DynamicFusion~\cite{newcombe2015dynamicfusion}. Recently, 3DGS has been integrated into several methods to enhance optimization speed and robustness. Examples include MoSca~\cite{lei2024mosca}, Dg-mesh~\cite{liu2024dynamic}, MoGS~\cite{ma2024reconstructing}, DynaSurfGS~\cite{cai2024dynasurfgs}, and Shape-of-Motion~\cite{wang2024shape}. Specifically, Dg-mesh~\cite{liu2024dynamic} introduces Gaussian-mesh anchoring to ensure Gaussians are evenly distributed, tracking mesh vertices over time, and producing high-quality meshes using a differential Poisson solver~\cite{peng2021shape}. Shape-of-Motion~\cite{wang2024shape} employs data-driven priors, such as monocular depth maps and 2D tracks, to constrain Gaussian motion, while DynaSurfGS~\cite{cai2024dynasurfgs} combines Gaussian features from 4D neural voxels with planar-based splatting for high-quality rendering and surface reconstruction. Despite progress, a fidelity gap persists due to the explicit regularization of Gaussian primitives, which restricts rendering quality relative to D3DGS~\cite{yang2024deformable}. Our approach addresses this with surface-aware density control of Gaussian, improving 3D reconstruction while maintaining high fidelity.

%% file: sec/2_xpreliminary.tex
\section{Preliminary}
\label{sec:pre}
\begin{figure*}
    \centering
    \includegraphics[width=0.85\linewidth]{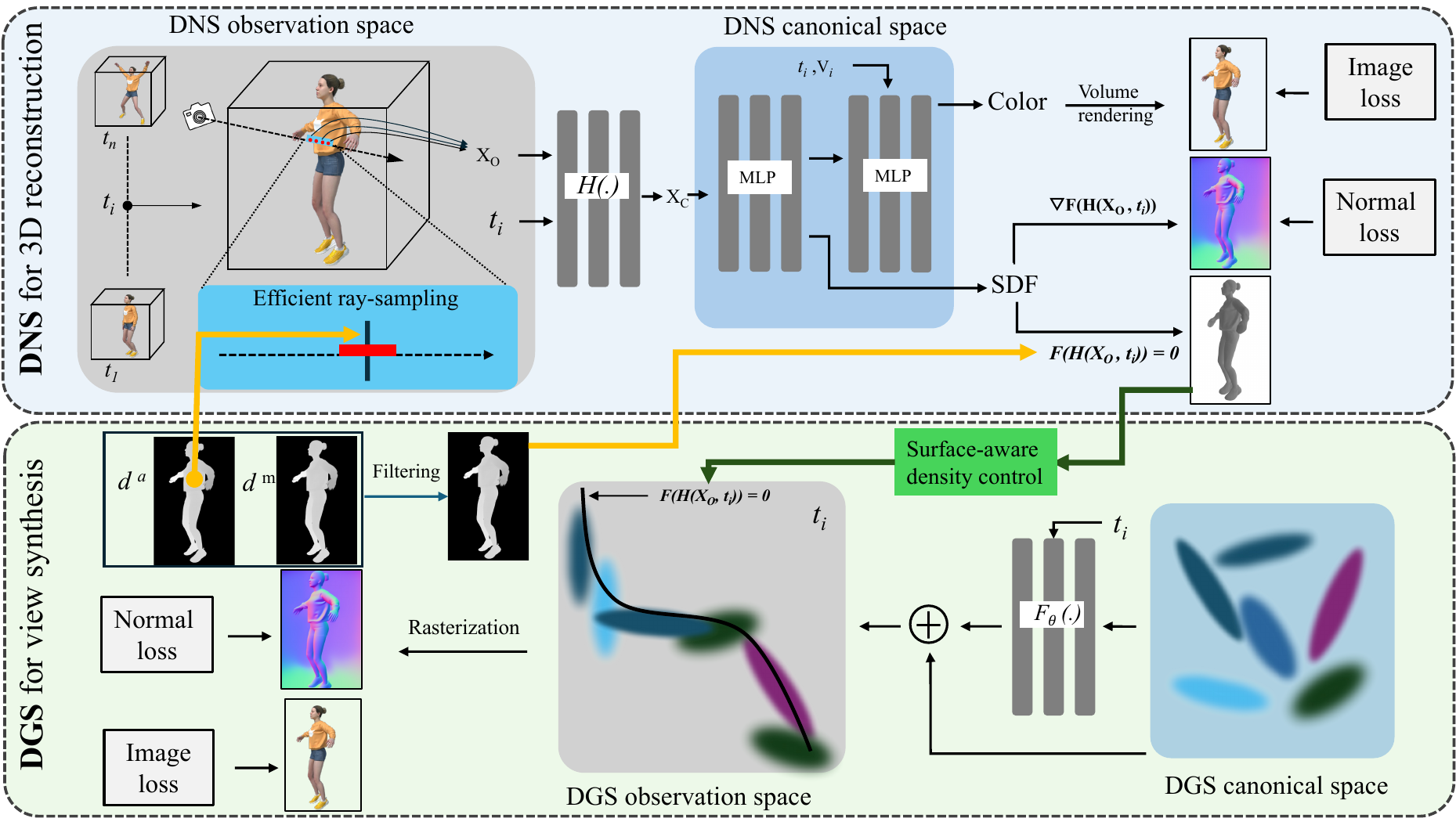}
    \label{fig:framewok}  
    \vspace{-1.0em}
    \caption{The framework of DGNS consists of two primary modules: the top module, DNS, for 3D reconstruction, and the bottom module, DGS, for view synthesis. There are three key interactions between these modules. The \textcolor{orange}{orange arrows} represent the information flow from DGS to DNS, including efficient ray-sampling and filtered depth supervision. The \textcolor{teal}{green arrow} illustrates the information flow from DNS to DGS, implementing surface-aware density control.}
    \vspace{-1.0em}
    \label{fig:framewok}
\end{figure*}

\subsection{Deformable 3DGS}
3DGS \cite{kerbl20233d} uses Gaussian primitives to represent a static scene, achieving high-quality rendering fidelity. Each Gaussian primitive is defined with a center ($x_i$), a covariance matrix \(\Sigma_i\), an opacity \(\sigma_i\), and spherical harmonics coefficients \(h_i\), i.e. \(G_i=\{{x_i},\Sigma_i,\sigma_i,h_i\}\). When rendering novel-view images, 3D Gaussian primitives are projected onto the 2D image plane and combined using $\alpha$-blending through a tile-based differentiable rasterizer. The color C(p) of a pixel p is computed as follows:
\vspace{-1.5em}

\begin{equation}
\begin{aligned}
C(p) &= \sum_{i \in N}  c_i \alpha_i \prod_{j=1}^{i-1} (1 - \alpha_j^{2D}), \\ 
\alpha_i &= \sigma_i e^{-\frac{1}{2} (p - \boldsymbol{x}_i)^T \Sigma{\prime}_i (p - \boldsymbol{x}_i)}
\end{aligned}
\label{equ:color_rendering}
\end{equation}

where $\Sigma{\prime}_i$ is the 2D projection of the 3D Gaussian’s covariance matrix. The rasterization process accumulates each Gaussian contribution efficiently, enabling high-quality rendering. However, standard 3DGS is limited to static scenes and cannot model temporal changes. To extend 3DGS to dynamic scenes, D3DGS~\cite{yang2024deformable} incorporates a deformation field that models time-dependent changes in position, rotation, and scale. Each Gaussian in canonical space is dynamically transformed by applying offsets calculated through a deformation MLP network $F_\theta$, \textit{i.e.},
$(\delta \mathbf{x}, \delta \mathbf{r}, \delta \mathbf{s}) = F_\theta(\gamma(\mathbf{x}), \gamma(t)),$ $F_\theta(.)$, where  $\gamma(\cdot)$  denotes positional encoding, $\mathbf{x}$ is the Gaussian’s canonical position, and $t$ is the current time step. The deformed Gaussian at time $t$ is expressed as: $G(\mathbf{x} + \delta \mathbf{x}, \mathbf{r} + \delta \mathbf{r}, \mathbf{s} + \delta \mathbf{s}, \sigma).$ The rasterization pipeline remains differentiable, allowing gradients to backpropagate through the Gaussian parameters and the deformation network during optimization. This framework handles both temporal consistency and fine-grained motion, achieving high-quality rendering.

\subsection{Dynamic Neural SDF}
SDF represents the object’s geometry by learning the signed distance of each point in a 3D space relative to the object's surface. Formally, the surface $S$ of an object is represented as the zero-level set of the SDF, defined by:
\begin{equation}
S = \{x \in \mathbb{R}^3 \mid \mathcal{F}(x) = 0\}    
\end{equation}
where $d(x)$ is the SDF value at a point $x$. Opacity used in 3DGS or NeRF can be derived from SDF value using a logistic function~\cite{wang2021neus}. Neural Dynamic Reconstruction (NDR)~\cite{cai2022neural} extends SDF to model dynamic objects by incorporating a deformation field. The deformation field provides a homeomorphic (continuous and bijective) mapping $\mathcal{H}(.,t)$: $\mathbb{R}^3 \rightarrow \mathbb{R}^3$, which maps $x_o$ of deformable observation space at time $t$ to its corresponding point $x_c$ in a canonical 3D space, where the SDF is defined independently of time or motion. This formulation ensures that any point on the dynamic surface is described accurately by setting  $d(x_c) = 0$. The Dynamic Neural Surface can be defined by:
\begin{equation}
ds = \{x \in \mathbb{R}^3 \mid \mathcal{F}(x_c) = \mathcal{F}(\mathcal{H}(x_o, t)) = 0\}    
\label{equ:biject_map}
\end{equation}
where the deformation field $\mathcal{H}(.)$ is designed strictly invertible, allowing a point $x_o$ in canonical space to map back to any observed frame $t$ via the inverse transformation $H^{-1}$. Invertibility of $\mathcal{H}$ enforces a cycle-consistent constraint across frames, which is a regularization for modeling dynamic scenes~\cite{wang2019learning}.

%% file: sec/3_methods.tex
\section{Method}
\label{sec:method}
Current approaches for dynamic scene reconstruction typically excel in either 3D geometry reconstruction or novel-view synthesis, but not both. Our framework, illustrated in Fig.~\ref{fig:framewok}, introduces a hybrid representation that combines Deformable Gaussian Splatting and Dynamic Neural Surfaces modules, with both components jointly optimized and mutually benefited to enhance performance across both tasks. The details of each module are presented in the following sections.

\subsection{DNS for Dynamic Surface Reconstruction}
\label{sec:method_ns}
The DNS module utilizes the deformation field defined in Eq.~(\ref{equ:biject_map}) to map the dynamic observation space to canonical space. Unlike other dynamic SDF methods\cite{cai2022neural}, our approach incorporates ray-sampling with depth proposals and depth supervision from the DGS module (as shown with \textcolor{orange}{orange arrows} in~\cref{fig:framewok}). Additionally, our method introduces normal supervision using surface normals generated by a foundation model.

\vspace{0.5em}
\noindent \textbf{Efficient Ray-sampling.}
Ray-sampling can be computationally intensive without prior information~\cite{mildenhall2021nerf}, as exhaustive sampling is needed to approximate a pixel’s color from an unknown density distribution accurately. Efficiency can be improved by selectively sampling in non-empty regions of the scene. NeRF’s hierarchical volume sampling scheme uses a coarse model to approximate a density distribution, guiding the sampling of the fine model and increasing computational efficiency. In addition to hierarchical sampling, two other major sampling schemes are proposal-based~\cite{Barron2022MipNeRF360} and occupancy-based~\cite{Yu2021PlenOctrees, Hu2022efficientnerf}. Proposal-based methods, like Mip-NeRF 360~\cite{Barron2022MipNeRF360}, replace the coarse model with a compact proposal model that produces only density rather than both density and color. Occupancy-based methods, such as PlenOctree~\cite{Yu2021PlenOctrees}, effectively filter out points with a low density, avoiding unnecessary sampling. While these coarse-to-fine strategies~\cite{mildenhall2021nerf, wang2021neus, rosu2023permutosdf, Barron2022MipNeRF360} improve efficiency, they often require costly rendering processes. In our framework, we use the depth map generated by the DGS branch to eliminate unnecessary queries in empty or occluded regions, thereby speeding up the ray-sampling process. The DGS depth map provides proximity to the surface, constraining the sampling range. Specifically, the $\alpha$-blending depth map $d^{\alpha}$ defines these sampling boundaries, and $d^{\alpha}$ is calculated as follows:
\begin{equation}
\scalebox{1.0}{$ d^{\alpha} = \sum_{i \in N} d_i \alpha_i \prod_{j=1}^{i-1} (1 - \alpha_j)/\sum_{i \in N} \alpha_i \prod_{j=1}^{i-1} (1 - \alpha_j)$}
\end{equation}
\noindent  where $N$ represents the count of 3D Gaussians encountered, $d_i$ is the distance from the $i$-th Gaussian to the camera, and $\alpha_i$ denotes opacity. The sampling process starts by emitting a ray from the camera center, $\vec{o}$, along a direction $\vec{v}$.  The ray-sampling points in observation space are taken near $\vec{o} + d^{\alpha} \cdot \vec{v}$, with the range adjusted based on SDF values calculated at ray-sampling points, \scalebox{0.9}{$ d = \mathcal{F}(\mathcal{H}(\vec{o} + d^{\alpha}  \cdot \vec{v}, t))$}, where $\mathcal{F}$ is to predict the SDF value for a point in observation space. The sampling interval along the ray ($\vec{o}$, $\vec{v}$) in observation space is from $\vec{o} + (d^{\alpha} - s|d|) \cdot \vec{v}$ to $\vec{o} + (d^{\alpha} + s|d|) \cdot \vec{v}$, in which the $s$ is a scaling factor for the predicted SDF value $d$. 

\begin{figure}
    \centering
    \includegraphics[trim=0 20 0 0, clip,width=1.0\linewidth]{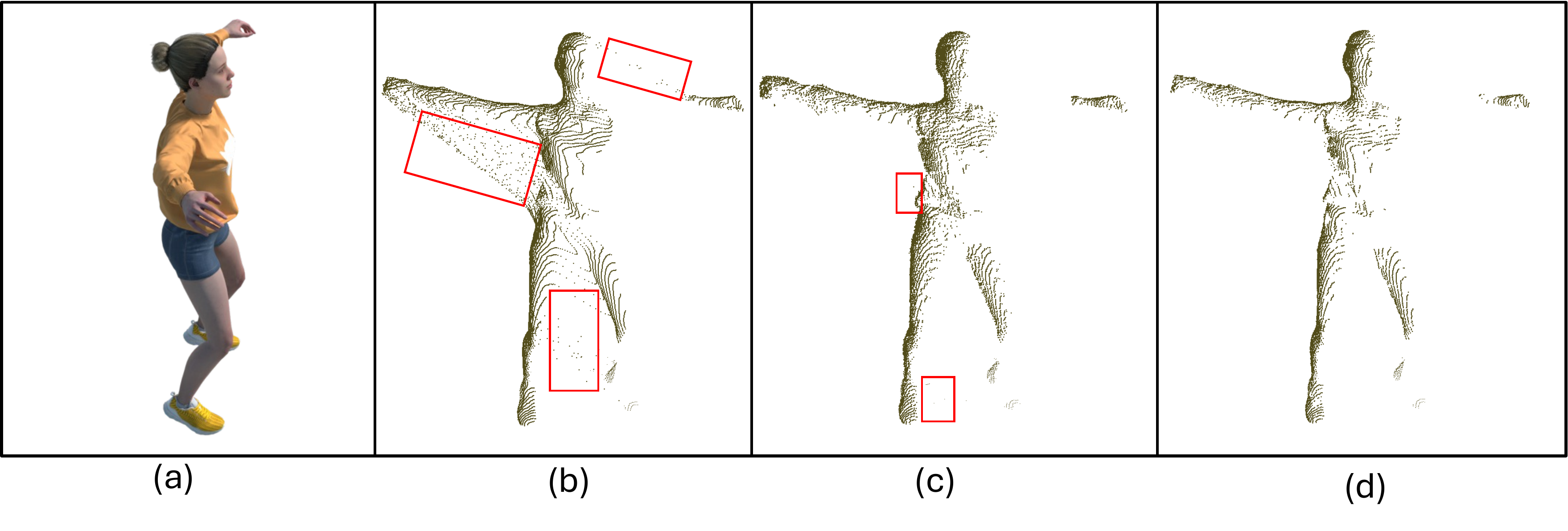}
    \vspace{-2.0em}
    \caption{Visualization of the depth map in 3D space. The leftmost is the RGB image, and images from left to right are 3D point clouds projected from $\alpha$-blending depth, median depth, and filtered depth.}
    \vspace{-1.5em}
    \label{fig:depth_maps}
\end{figure}

\vspace{0.5em}
\noindent \textbf{Depth and Normal Regularization.}
Recovering dynamic 3D structures from monocular video is a highly under-constrained optimization problem. Additional depth cues, such as RGB-D sensors or monocular depth estimation, are usually introduced to improve the reconstruction~\cite{lin2022occlusionfusion, cai2022neural, newcombe2015dynamicfusion}. The efficient ray-sampling mechanism can provide a rough sampling range but does not directly optimize the SDF in canonical space. The $\alpha$-blending depth tends to be noisy, especially in the edge regions, as shown in Fig.~\ref{fig:depth_maps} (b). To avoid the depth floaters around depth boundaries, the median depth is calculated simultaneously with Gaussian rasterization. The median depth of a ray is the depth of the Gaussian center which causes the accumulated rays' transmittance to drop below a threshold $\tau_{\text{d}}$. Therefore, $d^{\text{m}}=d_k\hspace{0.5em}\text{if}\hspace{0.25em}\scalebox{0.85}{$T_{k-1}$}  \geq \tau_{\text{d}}\hspace{0.25em}\text{and}\hspace{0.25em}\scalebox{0.85}{$T_k$} < \tau_{\text{d}}\hspace{0.25em} $, where \scalebox{0.85}{$T_i = \prod_{j=1}^{i-1} (1 - \alpha_j)$}, and $\tau_{\text{d}}$ is set to 0.6 in our experiment. The median depth is visualized in Fig.~\ref{fig:depth_maps} (c), and we can observe that there are still floaters around the surface due to the transmittance drop. To generate an accurate depth map for supervising the DNS, we propose a simple but effective filtering process, in which the depth is treated as a reliable prediction as long as the median depth and $\alpha$-blending one are close enough (i.e., smaller than $\tau_{\text{f}}$), as follows:
\begin{equation}
\scalebox{1.0}{$ d^f = 
\begin{cases}
( d^{\alpha} - d^m )/2, & \text{if } \left| d^{\alpha} - d^m \right| < \tau_{\text{f}} \\
0, & \text{if } \left| d^{\alpha} - d^m \right| \geq \tau_{\text{f}}
\end{cases}    $}
\label{equ:filtered_depth}
\end{equation}
where filtered points are on the surface with SDF loss, as follows:
\begin{equation}
\scalebox{1.0}{$  \mathcal{L}_{\text{sdf}} = \sum_{{d}^f \in \mathcal{D}} \| \mathcal{F}(\mathcal{H}(\vec{o} + {d}^f \cdot \vec{v}, t)) \|_1 $}
    \label{equ:sdf} 
\end{equation}
Apart from depth cues, we introduce the foundation model to provide additional normal supervision to relieve the under-constrained optimization problem. The monocular normal foundation models, i.e. Marigold~\cite{ke2023repurposing, martingarcia2024diffusione2eft}, are used to generate the pseudo normal map \scalebox{0.9}{$\bar{{N}}$}. Similar to MonoSDF~\cite{tian2023mononerf}, we use the volume rendering method to calculate the normal of ray $\mathbf{r}$, denoted as \scalebox{0.9}{$\hat{{N}}$}, which is a weighted sum of each ray-sampling point normal, \scalebox{0.9}{$\nabla_{\mathbf{x_o}} \mathcal{F}(\mathcal{H}(x_o, t)))$}, on the ray $\mathbf{r}$. The consistency between the volume rendered normal map \scalebox{0.9}{$\hat{{N}}$} and the predicted monocular normal map \scalebox{0.9}{$\bar{{N}}$} is imposed with angular and $L1$ loss~\cite{tian2023mononerf, zhang2024pmvc}, as follows:
\begin{equation}
\scalebox{1.0}{$ \mathcal{L}_{\text{dn}}^{n} = \frac{1}{N}\sum_{\mathbf{r} \in \mathcal{R}} \left\| \hat{{N}}(\mathbf{r}) - \bar{{N}}(\mathbf{r}) \right\|_1 + \left\| 1 - \hat{{N}}(\mathbf{r})^{\top} \bar{{N}}(\mathbf{r}) \right\|_1 $}
\end{equation}

\subsection{DGS for Dynamic View Synthesis}
\label{sec:method_gs}

In contrast to existing methods using the 3DGS to model dynamic scenes~\cite{yang2024deformable, liu2024dynamic, yang2023real, kratimenos2023dynmf}, we add the surface-aware density control with the geometry guidance from DNS module (as shown with \textcolor{teal}{green arrow} in~\cref{fig:framewok}),  concentrating deformable 3D Gaussian points around the surface area, which enhances the model’s ability to capture both the geometry and color of dynamic surfaces. Additionally, we incorporate surface normal supervision using normals derived from foundation models. These components are detailed below.

\vspace{0.5em} 
\noindent \textbf{Surface-aware Density Control.}
Utilizing the object’s surface can serve as effective guidance for positioning Gaussian primitives to enhance rendering quality~\cite{wang2023adaptive, yu2024gsdf, lu2024scaffold}. However, directly aligning Gaussian primitives to the surface often causes a decline in rendering quality~\cite{zhang2024neural, guedon2024sugar}. To address this, we adopt a surface-aware density control strategy, similar to GSDF~\cite{yu2024gsdf} for static scenes, to optimize the distribution of Gaussian primitives. Specifically, the zero-level set of DNS (see Eq.~(\ref{equ:biject_map})) in observation space is used to guide Gaussian growth (split/clone) and pruning operations. Gradient-based adaptive density control\cite{kerbl20233d} and the SDF values of Gaussian primitives from the DNS module are employed to fine-tune Gaussian placement and density. For each Gaussian primitive $\mathbf{x}_{g}$ in DGS canonical space, we determine its location in observation space as ($\mathbf{x}_{g}+\delta \mathbf{x}_{g}$), where $(\delta\mathbf{x}_{g}, \delta\mathbf{r}_{g}, \delta\mathbf{s}_{g}) = F_\theta(\gamma(\mathbf{x}_{g}), \gamma(t))$  through the GS deformation field. The SDF distance of $\mathbf{x}_{g}$ is then calculated as ${d}_{g} = \mathcal{F}(\mathcal{H}(\mathbf{x}{g} + \delta \mathbf{x}_{g}, t))$ . Accordingly, the criteria for Gaussian growth are defined as follows:
\vspace{-0.8em}
\begin{equation}
\epsilon_{g} = \nabla_{\mathbf{x}_{g}} + w_{g} \phi(d_{g}),
\end{equation}

where $\nabla_{\mathbf{x}_{g}}$ represents the average gradient of $\mathbf{x}_{g}$, $w_{g}$ is a weighting parameter which controls the influence of geometric factors, and $\phi(x)=\exp\left(-{x^2}/{2\sigma^2}\right)$ is inversely proportional to the SDF value. When $\epsilon_{g}$ is larger than a threshold $\tau_{\text{g}}$, the new Gaussians will be added. In addition to adding Gaussians, the SDF distance can be used to prune Gaussian primitives that lie far from the surface. The pruning criteria are customized as follows:
\vspace{-0.8em}
\begin{equation}
\epsilon_p = \sigma_p - w_p (1 - \phi(d_{g})),
\end{equation}

where $\sigma_p$ is the sum of opacity across $K$ iterations and the weighting parameter $w_p$ controls the influence of the SDF value. Gaussian primitives will be removed if $\epsilon^p$ is below a set threshold $\tau^p$.

\vspace{0.5em}
\noindent \textbf{Normal Supervision.}
To ease the under-constrained optimization problem of monocular dynamic scene reconstruction, we add the normal loss to regularize the GS module. The normal direction of each Gaussian primitive can be approximated with the direction of the axis with the minimum scaling factor~\cite{chen2023neusg}. The normal in the world coordinate system is defined as: $ n= \mathbf{R}[k, :] \in \mathbb{R}^3,\hspace{0.25em}k = \arg \min ([s_1, s_2, s_3])$, where $s_1, s_2, s_3$ are the Gaussian scales and $\mathbf{R}$ is the Gaussian rotation matrix. Similar to rendering color in~\cref{equ:color_rendering}, the normal vector of a point $p$ in the screen space can be rendered as \scalebox{0.9}{$\hat{N}^{g}(p) = \sum_{i \in N} n_i \alpha_i \prod_{j=1}^{i-1} (1 - \alpha_j)$}. Therefore, the normal regularization loss can be calculated as:
\vspace{-0.8em}
\begin{equation}
\scalebox{1.0}{$ \mathcal{L}_{\text{dg}}^{\text{n}} =\frac{1}{N} \sum_{p \in \mathcal{P}} \left\| \hat{N}^g(p) - \bar{N}(p) \right\|_1 + \left\| 1 - \hat{{N}}^g(p)^{\top} \bar{{N}}(p) \right\|_1 $}
\end{equation}
\vspace{-0.5em}
\begin{figure*}
    \centering
    \includegraphics[trim=30 40 30 0, clip,width=0.9\linewidth]{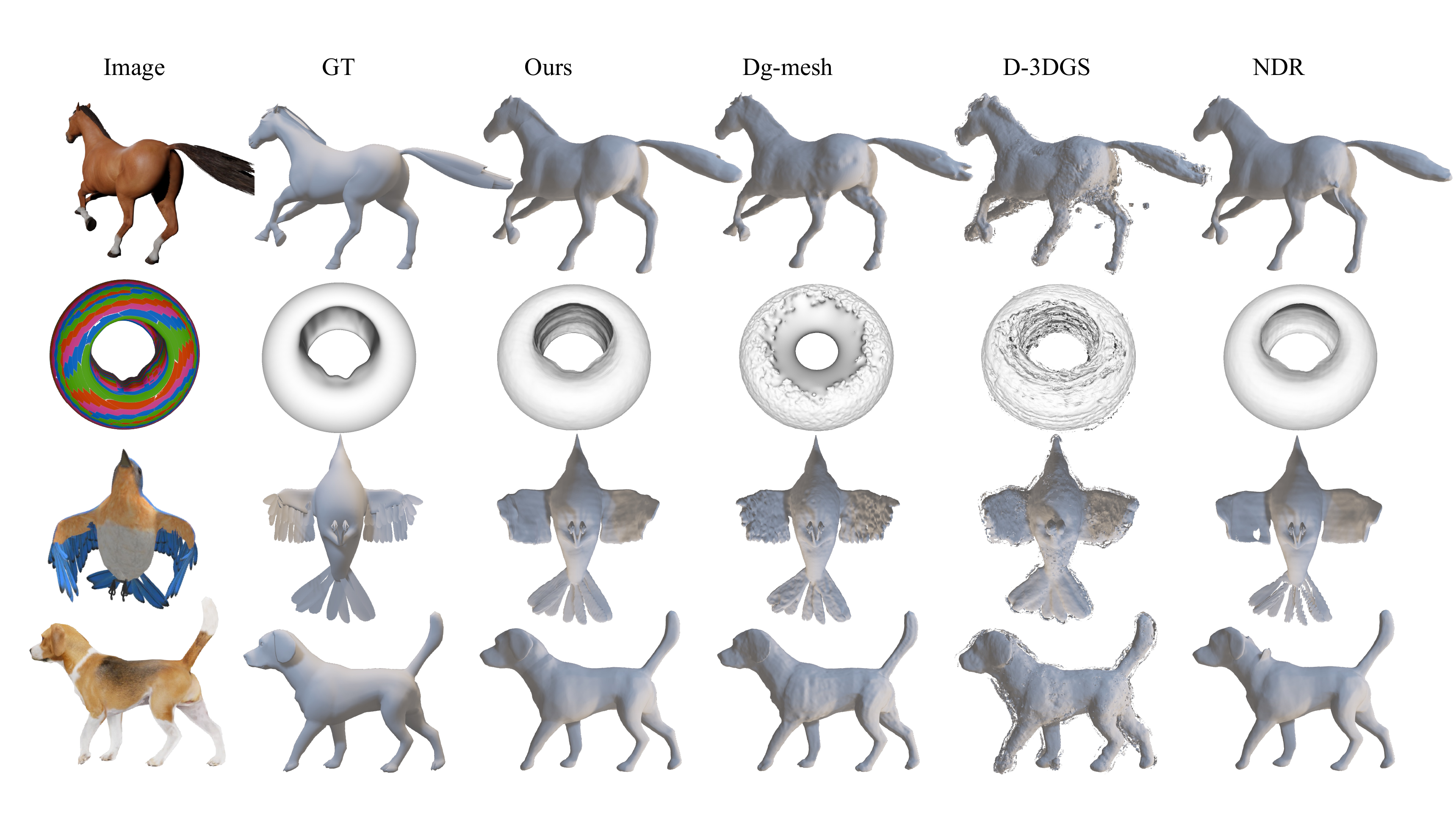}
    \vspace{-2em}
    \caption{Qualitative comparison on the Dg-mesh dataset. The samples, from top to bottom, are \textit{Horse}, \textit{Torus2sphere}, \textit{Bird}, and \textit{Beagle}. Compared to other baselines, our results are the closest to the ground truth (GT).}
    \vspace{-1em}
    \label{fig:dg-mesh-vis}
\end{figure*}

\subsection{Optimization}
For the GS module, the image loss measures the difference between the rendered RGB images and ground truth images. Usually, it includes two rendering losses $\mathcal{L}_1$ and $\mathcal{L}_{\text{ssim}}$, supplemented by a normal loss:
\vspace{-0.8em}
\begin{equation}
   \scalebox{1.0}{$\mathcal{L}_{\text{dg}} = \lambda_I \mathcal{L}_1 + (1 - \lambda_I) \mathcal{L}_{\text{ssim}} + \lambda_{\text{gn}} \mathcal{L}_{\text{dg}}^{\text{n}}$}
\end{equation}
where $\lambda_I$ and $\lambda_{\text{gn}}$ are the weighting coefficients. Similarly, the image loss in the DNS module is supervised by the $\mathcal{L}_1$ loss. The SDF will be regularized by Eikonal loss $\mathcal{L}_{eik}$~\cite{gropp2020implicit}. Moreover, the SDF is supervised by the filtered point which generated the $\mathcal{L}_{\text{sdf}}$ as Eq.~(\ref{equ:sdf}). Thus, the loss is designed as:
\vspace{-0.5em}
\begin{equation}
\scalebox{1.0}{$\mathcal{L}_{\text{dn}} = \mathcal{L}_1 + \lambda_{\text{sdf}} \mathcal{L}_{\text{sdf}} + \lambda_{\text{nn}} \mathcal{L}_{\text{dn}}^{n} + \lambda_{\text{eik}} \mathcal{L}_{\text{eik}}$}   
\end{equation}
where $\lambda_{\text{sdf}}$, $\lambda_{\text{nn}}$, and $\lambda_{\text{eik}}$ are the weighting parameters for each loss term. The final total loss will be: 
\vspace{-0.5em}
\begin{equation}
\scalebox{1.0}{$\mathcal{L} =  \mathcal{L}_{dg} + \mathcal{L}_{dn} $} 
\end{equation}

With the $\mathcal{L}$, the proposed DGNS learns hybrid representations jointly across two tasks (\textit{i.e.}, 3D geometry reconstruction and novel view synthesis) using a unified framework, and two modules mutually benefit from each other through efficient ray-sampling, depth regularization, and guided density control.

%% file: sec/4_experiments.tex
\section{Experiments}
\label{sec:exp}
\label{sec:experiments}
\subsection{Setup}

\begin{table*}[ht]
    \centering
    \caption{Mesh reconstruction and rendering quality results of our method compared to other baselines on Dg-mesh~\cite{liu2024dynamic}. Reconstructed meshes are measured with Chamfer Distance (CD) and Earth Mover Distance (EMD) with the ground truth mesh. Rendering quality is measured with Peak Signal-to-Noise Ratio (PSNR). The color of each cell indicates the \colorbox{best}{best}, \colorbox{second}{second}, and \colorbox{third}{third} scores, and the third-best results. In general, our method produces a better reconstruction and rendering quality.}  
    \vspace{-1.2em}
\begin{adjustbox}{width=0.8\textwidth, height=4.5cm}
    \begin{tabular}{c|ccc|ccc|ccc}
        \toprule[1.5pt]
        \multirow{2}{*}{Method} & \multicolumn{3}{c|}{Duck} & \multicolumn{3}{c|}{Horse} & \multicolumn{3}{c}{Bird} \\
        & CD $\downarrow$ & EMD $\downarrow$ & PSNR $\uparrow$ & CD $\downarrow$ & EMD $\downarrow$ & PSNR $\uparrow$ & CD $\downarrow$ & EMD $\downarrow$ & PSNR $\uparrow$ \\
        \midrule[0.7pt]
        D-NeRF  &\cellcolor{third} 0.934 & 0.073       & 29.79 & 1.685                  & 0.280                   & 25.47 & 1.532 & 0.163 & 23.85 \\
        NDR     &2.235                   & 0.083       & 24.79 &\cellcolor{third} 0.359 & \cellcolor{second}0.148 & 27.23 & \cellcolor{second} 0.444 & \cellcolor{third}0.130  & 25.43 \\
        K-Plane & 1.085 & \cellcolor{third}0.055 & 33.36 & 1.480 & 0.239 & 28.11 & 0.742 &  0.131 & 23.72 \\
        HexPlane & 2.161 & 0.090 & 32.11 & 1.750 & 0.199 & 26.78 & 4.158 & 0.178 & 22.19 \\
        TiNeuVox & 0.969 & 0.059 &  \cellcolor{third}34.33 & 1.918 & 0.246 & \cellcolor{third} 28.16 & 8.264 & 0.215 & \cellcolor{third}25.55 \\
        D3DGS &1.643 & 0.1146 &\cellcolor{second}36.76 &0.941 &0.203 &\cellcolor{best}35.66 &1.398 & 0.138 & \cellcolor{second} 29.38 \\
        Dg-mesh & \cellcolor{second}0.790 & \cellcolor{second}0.047 & 32.89 & \cellcolor{second}0.299 & \cellcolor{third}0.168 & 27.10 & \cellcolor{third}0.557 & \cellcolor{second}0.128 & 22.98 \\    
        \midrule[0.4pt]
        DGNS (ours) & \cellcolor{best}0.773 & \cellcolor{best}0.046 & \cellcolor{best}37.39 & \cellcolor{best}0.289 & \cellcolor{best}0.142 & \cellcolor{second}35.52 & \cellcolor{best}0.440 & \cellcolor{best}0.126 & \cellcolor{best}30.23 \\
        \midrule[1.5pt]
        \multirow{2}{*}{Method} & \multicolumn{3}{c|}{Beagle} & \multicolumn{3}{c|}{Torus2sphere} & \multicolumn{3}{c}{Girlwalk} \\
        & CD $\downarrow$ & EMD $\downarrow$ & PSNR $\uparrow$ & CD $\downarrow$ & EMD $\downarrow$ & PSNR $\uparrow$ & CD $\downarrow$ & EMD $\downarrow$ & PSNR $\uparrow$ \\
        \midrule[0.5pt]
        D-NeRF & 1.001 & 0.149 & 34.47 & \cellcolor{third}1.760 & 0.250 & 24.23 & 0.601 & 0.190 & 28.63 \\
        NDR    & \cellcolor{third}0.747  & \cellcolor{second} 0.106  &28.39 & 1.767 & 0.176 & 25.19 & \cellcolor{second}0.482 & 0.158 & 29.98\\
        K-Plane & 0.810 &  0.122 & 38.33 & 1.793 &  \cellcolor{third}0.161 & \cellcolor{best}31.22 & \cellcolor{third}0.495 & 0.173  &  32.12 \\
        HexPlane & 0.870 & 0.115 & 38.03 & 2.190 & 0.190 & 29.71 & 0.597 & \cellcolor{third}0.155 & 31.77 \\
        TiNeuVox & 0.874 & 0.129 &\cellcolor{second} 38.97 & 2.115 & 0.203 & 28.76 & 0.568 & 0.184 & \cellcolor{third} 32.81 \\
        D3DGS & 0.898 & 0.211 & \cellcolor{third} 38.56 & 1.487 & 0.193 & \cellcolor{second}30.22 & 0.883 & 0.270 & \cellcolor{best} 38.75 \\
        Dg-mesh & \cellcolor{best}0.639 & \cellcolor{third}0.117 & 34.53 & \cellcolor{second}1.607 & \cellcolor{second}0.172 & 26.61 &  0.726 & \cellcolor{second}0.136 & 28.64 \\
        \midrule[0.4pt]
        DGNS (ours) & \cellcolor{second}0.643 & \cellcolor{best}0.101 &\cellcolor{best} 39.29 & \cellcolor{best}0.456 & \cellcolor{best}0.153 & \cellcolor{third} 29.76 & \cellcolor{best}0.413 & \cellcolor{best}0.124 & \cellcolor{second} 38.64 \\
        \bottomrule[1.5pt]
    \end{tabular}
\end{adjustbox}   
\vspace{-1em}
    \label{tab:dg_mesh_dataset}
\end{table*}

\begin{figure}[t]
    \centering
    \includegraphics[trim=0 80 0 0, clip, width=1.0\linewidth]{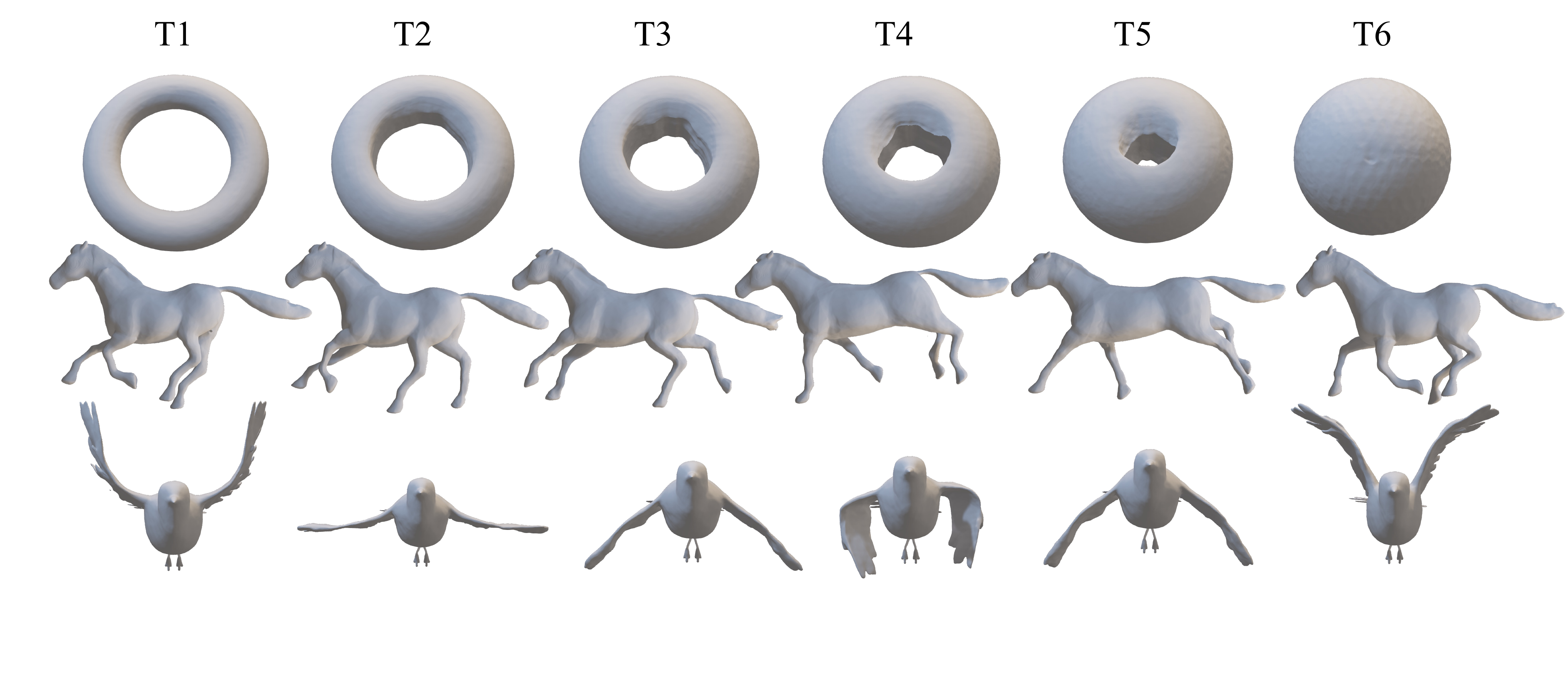}
    \vspace{-3em}
    \caption{Qualitative results demonstrating the temporal evolution of 3D meshes reconstructed by our method. Rows (from top to bottom) show the \textit{Torus2sphere}, \textit{Horse}, and \textit{Bird} sequences from the dg-mesh dataset. Columns (from left to right) depict different time points}
    \vspace{-3em}
    \label{fig:dynamic_change}
\end{figure}

\noindent \textbf{Dataset and Baseline.}
In this work, we conducted experimental evaluations on three public monocular video datasets: two synthetic datasets, D-NeRF~\cite{pumarola2021d} and Dg-mesh~\cite{liu2024dynamic}, and one real dataset, Nerfies~\cite{park2021nerfies}. Comprehensive quantitative analyses and selected qualitative results are presented in the main paper, while additional qualitative results can be found in the supplementary material. D-NeRF includes eight sets of dynamic scenes featuring complex motion, such as articulated objects and human actions. Dg-mesh provides six sets of dynamic scenes with ground truth for each object’s deformable 3D structure. Both datasets have images at $800\times800$ resolution, with $100$ to $200$ images per scene. To demonstrate the effectiveness of our method, we compared it with seven baselines: D-NeRF~\cite{pumarola2021d}, NDR~\cite{cai2022neural}, K-Plane~\cite{fridovich2023k}, HexPlane~\cite{cao2023hexplane}, TiNeuVox~\cite{fang2022fast}, Dg-mesh~\cite{liu2024dynamic}, and D3DGS~\cite{yang2024deformable}. D-NeRF, TiNeuVox, and NDR use implicit representations with a deformation field to map the dynamic observation space to static canonical space, while K-Plane and HexPlane employ 4D feature volumes with volume factorization. In contrast, Dg-mesh and D3DGS rely on explicit 3D Gaussian representations to model dynamic scenes. Our method integrates SDF and 3D Gaussians, achieving state-of-the-art performance in both 3D reconstruction and view synthesis.

\vspace{0.5em}
\noindent \textbf{Implementations.}

For the DNS module, we utilize the bijective mapping network~\cite{dinh2016density, cai2022neural} to learn the deformation field to mapping points from observation space back to canonical space, and canonical space is the hybrid of hash-grid encoders and 4-layer MLP for speedup. The entire training process includes a warm-up phase ($0$ to $10$k iterations) followed by joint training ($10$k to $40$k iterations). For efficient ray sampling, the scaling factor $s$ is set to $3$ from $10$k to $20$k iterations for coarse search, and then to $1$ for the remainder of the training. 
The DGS model uses an 8-layer MLP (256 channels) for deformation learning. DGS takes around $10$k iterations to make an accurate depth prediction, from which depth guidance for ray and SDF supervision in NS starts. After the $15$k iteration warm-up, the geometry guidance for density control in DGS begins from $15$k iterations. Joint optimization lasts for $25$k iterations. The entire training process includes a warm-up phase ($0$ to $15$k iterations) followed by joint training ($15$k to $40$k iterations). The weight $\lambda_I$ is set to $0.8$ for calculating image loss, while $\lambda_{gn}$ starts to take effect at $10$k iterations with a value of $0.1$. The deformable network activates after $3$k iterations, with density control starting at 500 iterations.
Gaustudio~\cite{ye2024gaustudio} extracts 3D meshes from dynamic Gaussian primitives~\cite{yang2024deformable}. All experiments used NVIDIA RTX A6000 GPU (48GB).

\subsection{Results}
\begin{table*}[ht]
\centering
\caption{Rendering quality results of our method compared with other baselines on D-NeRF~\cite{pumarola2021d}. Our method is on par with the state-of-the-art D3DGS, yet significantly outperforms other baseline methods.}
\vspace{-1em}
\begin{adjustbox}{width=0.9\textwidth, height=4.0cm}
    \begin{tabular}{c|ccc|ccc|ccc|ccc}
    \toprule[1.5pt]
    \multirow{2}{*}{Method} & \multicolumn{3}{c|}{Hell Warrior} & \multicolumn{3}{c|}{Mutant} & \multicolumn{3}{c|}{Hook} & \multicolumn{3}{c}{Bouncing Balls} \\
                            & PSNR ↑    & SSIM ↑    & LPIPS ↓   & PSNR ↑    & SSIM ↑    & LPIPS ↓   & PSNR ↑    & SSIM ↑    & LPIPS ↓   & PSNR ↑    & SSIM ↑    & LPIPS ↓   \\ \midrule[0.7pt]
    D-NeRF                  & 24.06               & 0.944    & 0.071     & 30.31     & 0.967    &\cellcolor{third}0.039 & 29.02   & \cellcolor{third} 0.960    & 0.055    & 38.17     & 0.989    & \cellcolor{third} 0.032    \\
    NDR                     &\cellcolor{third}31.58 &0.971     &\cellcolor{third} 0.035      & 29.83     &0.958	   &0.049    & 26.62   &0.952     &0.071 & 36.34	& 0.986 &0.039\\
    TiNeuVox                & 27.10                & 0.964    & 0.077     & 31.87     & 0.961    & 0.047    &\cellcolor{third}30.61& \cellcolor{third} 0.960    & 0.059    & \cellcolor{third} 40.23     & \cellcolor{third} 0.993    & 0.044    \\
    Tensor4D                & 31.26                 &\cellcolor{third}0.975    &\cellcolor{second}0.024& 29.11     & 0.965    & 0.056    & 24.47   & 0.943    & 0.064    & 26.56     & 0.992    & 0.044    \\
    K-Planes                & 24.58                 & 0.952    & 0.084     &\cellcolor{third}32.50     & 0.951    & 0.066    & 28.12   & 0.949    & \cellcolor{third} 0.052    & 40.05     & \cellcolor{third} 0.993    & \cellcolor{third} 0.032    \\
    D3DGS                   &\cellcolor{second} 41.54 &\cellcolor{second}0.987&\cellcolor{best}0.023     &\cellcolor{best}42.63&\cellcolor{best}0.995&\cellcolor{best}0.005&\cellcolor{best}37.42&\cellcolor{best}0.987&\cellcolor{best}0.014&\cellcolor{best}41.01&\cellcolor{best}0.995&\cellcolor{best}0.009    \\ 
    Dg-mesh                 & 25.46                 & 0.959    & 0.074     & 30.40     &\cellcolor{third}0.968&0.055&27.88& 0.954    & 0.074    & 29.15     & 0.969    & 0.099     \\  
    \midrule[0.4pt]
    DGNS (ours)          &\cellcolor{best}41.68 &\cellcolor{best}0.988&\cellcolor{best}0.023&\cellcolor{second}41.47 &\cellcolor{second}0.984&\cellcolor{second}0.011 &\cellcolor{second} 36.34&\cellcolor{second}0.979&\cellcolor{second}0.023&\cellcolor{second}40.71&\cellcolor{second}0.994 &\cellcolor{second}0.012 \\ 
    \midrule[1.5pt]
    \multirow{2}{*}{Method} & \multicolumn{3}{c|}{Lego}         & \multicolumn{3}{c|}{T-Rex}  & \multicolumn{3}{c|}{Stand Up} & \multicolumn{3}{c}{Jumping Jacks} \\
                            & PSNR ↑    & SSIM ↑    & LPIPS ↓   & PSNR ↑    & SSIM ↑    & LPIPS ↓   & PSNR ↑    & SSIM ↑    & LPIPS ↓   & PSNR ↑    & SSIM ↑    & LPIPS ↓   \\ \midrule[0.7pt]
    D-NeRF                  & 25.56     & 0.936    & 0.082    & 30.61     & 0.967    & 0.054    & 33.13     & 0.978    & 0.036    & 32.70     & 0.978    & 0.039    \\
    NDR  & 24.54 &0.936 &0.084 & 23.47&	0.963&0.047& 30.22 &0.961 &0.047 & 29.02&0.957&0.043\\
    TiNeuVox                & 26.64     & 0.953    & 0.078    & \cellcolor{third} 31.25     & 0.967    & 0.048    & \cellcolor{third}34.61     & \cellcolor{third} 0.980    & 0.033    & \cellcolor{third} 33.49     & 0.977    & \cellcolor{third} 0.041    \\
    Tensor4D                & 23.24     & 0.942    & 0.089    & 23.86     & 0.954    & 0.054    & 26.30     & 0.938    & 0.056    & 24.20     & 0.925    & 0.067    \\
    K-Planes                & \cellcolor{third} 28.91 & \cellcolor{third}  0.970    & 0.035    & 30.43     & \cellcolor{third} 0.970    &\cellcolor{third} 0.031 & 33.10     & 0.979    & \cellcolor{third} 0.031    & 31.11     & 0.971    & 0.047    \\
    D3DGS                   &\cellcolor{best}33.07&\cellcolor{best}0.979&\cellcolor{best}0.018&\cellcolor{best}38.10 &\cellcolor{best}0.993 &\cellcolor{best}0.010&\cellcolor{second}44.62&\cellcolor{second}0.995&\cellcolor{best}0.006&\cellcolor{second}37.72&\cellcolor{second}0.990 &\cellcolor{second}0.013    \\
    Dg-mesh                 & 21.29     & 0.838    & 0.159    & 28.95     & 0.959    & 0.065    & 30.21     & 0.974    & 0.051    & 31.77     &\cellcolor{third}0.980     & 0.045   \\ 
    \midrule[0.4pt]
    DGNS (ours)           &\cellcolor{second}29.48&\cellcolor{second}0.974&\cellcolor{second}0.026&\cellcolor{second}37.68 &\cellcolor{second}0.989 &\cellcolor{second}0.012&\cellcolor{best}44.81&\cellcolor{best}0.996&\cellcolor{second}0.007&\cellcolor{best}38.60&\cellcolor{best}0.993&\cellcolor{best}0.011 \\ 
    \bottomrule[1.5pt]
    \end{tabular}
\end{adjustbox}
\vspace{-1em}
\label{tab:d_nerf_dataset}
\end{table*}

\noindent \textbf{Dg-mesh Dataset.}
The Dg-mesh dataset~\cite{liu2024dynamic} provides geometrical ground truth for dynamic objects at each timeframe, enabling a thorough quantitative evaluation of our method, DGNS, for both 3D reconstruction and novel-view synthesis tasks across six objects. As shown in Tab.~\ref{tab:dg_mesh_dataset}, our method consistently achieves the lowest CD and EMD scores across nearly all object categories. This performance highlights DGNS’s ability to capture fine-grained dynamic spatial structures with higher accuracy than competing methods, while also surpassing baselines in rendering quality. Regarding reconstruction accuracy, Dg-mesh frequently ranks second, especially in achieving lower CD and EMD scores, however, this improvement comes at the cost of rendering quality. Conversely, while D3DGS exhibits competitive performance in novel-view synthesis, it fails to achieve the same low CD and EMD scores necessary for high-accuracy 3D reconstruction. Our method is unique in offering consistently superior performance across both 3D reconstruction and rendering quality, providing a comprehensive solution for tasks requiring both structural precision and visual fidelity.

The qualitative results are presented in Figs.~\ref{fig:dg-mesh-vis} and~\ref{fig:dynamic_change}. In Fig.~\ref{fig:dg-mesh-vis}, it is shown that meshes reconstructed by DGNS (ours) most closely resemble the ground truth. For example, Gaussian-based methods like D3DGS~\cite{yang2024deformable} struggle with surface accuracy due to floating Gaussian points, resulting in less cohesive 3D surfaces. While Dg-mesh~\cite{liu2024dynamic} improves upon this with an anchoring process to reduce floating points, the resultant surface still lacks sufficient detail and smoothness. The NDR~\cite{cai2022neural} method encounters challenges with specific object parts, such as accurately reconstructing the legs in \textit{Girlwalk}, the tails of \textit{Bird}, and the inner surface of \textit{Torus2sphere}. Fig.~\ref{fig:dynamic_change} demonstrates our method’s capability to effectively model dynamic samples exhibiting extreme deformations, rapid motion, or topological changes, which are critical for validating its robustness in real-world scenarios. Additional qualitative results are provided in the supplementary materials.

\begin{figure*}
    \centering
    \includegraphics[trim=0 40 0 0, clip, width=0.85\linewidth]{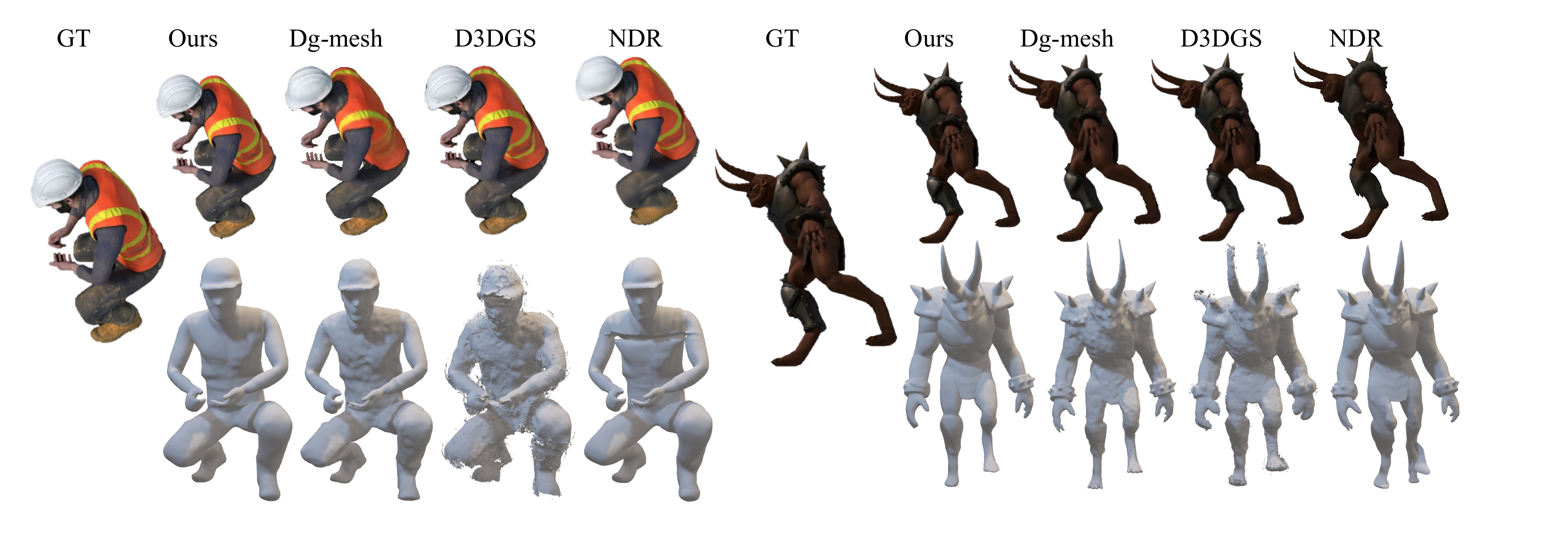}
    \vspace{-2em}
    \caption{Qualitative comparison on the D-NeRF dataset. The samples from right to left are \textit{Hellwarrior} and \textit{Standup}. Our method can achieve 3D reconstructions with smooth surfaces.}
    \vspace{-1em}
    \label{fig:synthesis_qualitative}
\end{figure*}

\vspace{0.5em}
\noindent \textbf{D-NeRF Dataset.}
The comparison of our method against baselines is presented in Tab.~\ref{tab:d_nerf_dataset}, demonstrating that our method, DGNS, provides a robust balance across PSNR, SSIM, and LPIPS metrics across various scenes. This balance underscores its effectiveness in novel-view synthesis tasks. While our method achieves comparable performance with D3DGS in rendering quality across some scenes, DGNS consistently surpasses other baselines, including Dg-mesh, the method previously noted for its efficacy in 3D reconstruction.

Notably, mesh ground truths are unavailable in the D-NeRF dataset. Thus, we have supplemented the quantitative analysis with a qualitative comparison, shown in Fig.~\ref{fig:synthesis_qualitative}. The qualitative assessment reveals that our method yields superior performance in 3D reconstruction fidelity, even when visually assessed against high-performing baselines. Although D3DGS achieves top scores in rendering quality, it does not maintain the structural accuracy in 3D reconstruction that our method consistently delivers. In contrast, our method demonstrates state-of-the-art rendering quality while excelling in structural reconstruction, presenting a compelling solution for applications requiring high fidelity in visual output and accurate 3D geometry. More qualitative results are included in the Supplementary material.

\begin{figure}[t]
    \centering
    \includegraphics[width=0.75\linewidth]{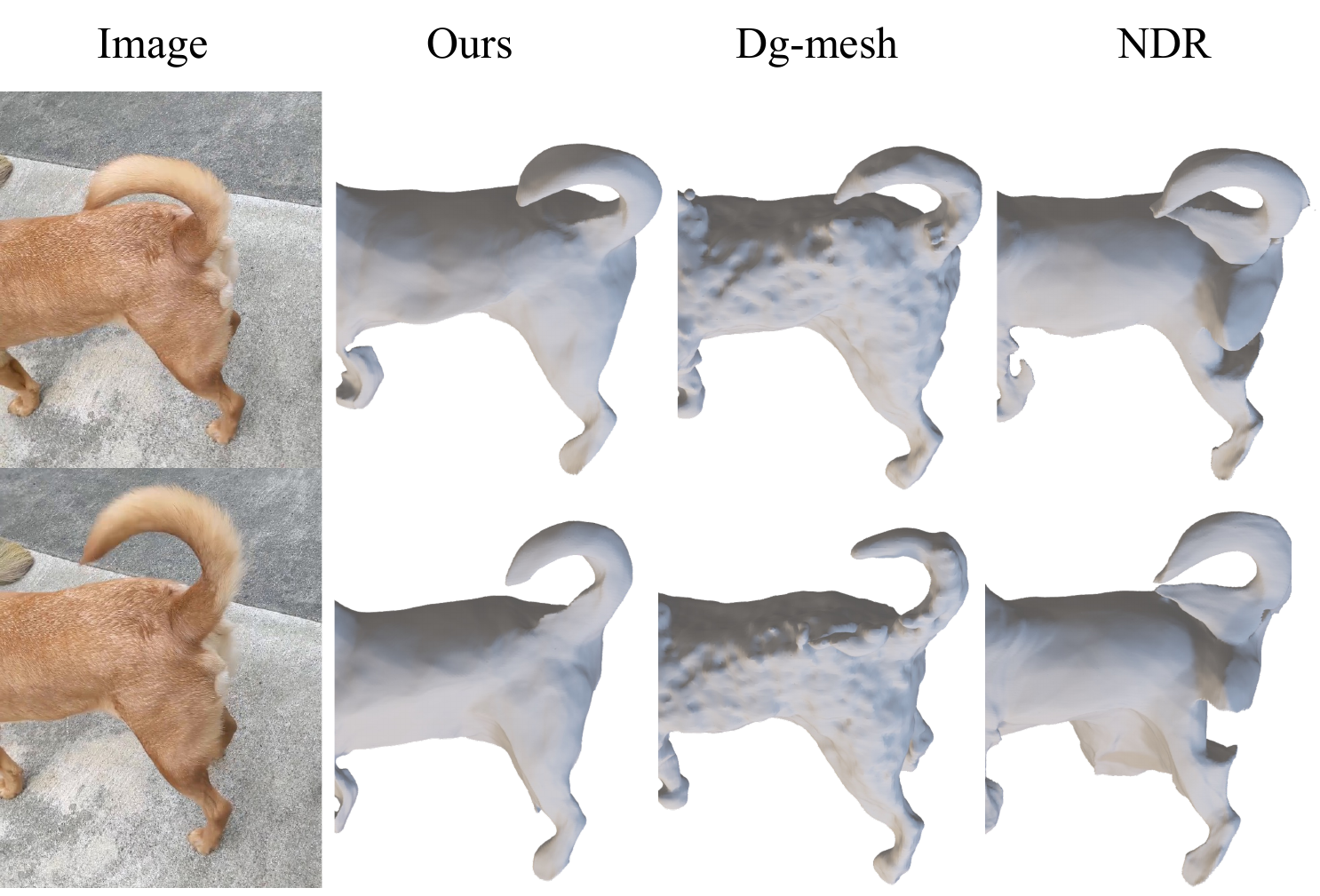}
    \vspace{-1.0em}
    \caption{Qualitative comparison on the Nerfies dataset.}
    \vspace{-1.5em}
    \label{fig:qualitative_comp_nerfies}
\end{figure}

\vspace{0.5em}
\noindent \textbf{Nerfies Dataset.}
Due to the absence of ground-truth data for the Nerfies dataset~\cite{park2021nerfies}, we provide a qualitative comparison, shown in Fig.~\ref{fig:qualitative_comp_nerfies}. As illustrated, our method produces smoother surfaces compared to DG-mesh and achieves more accurate geometric reconstructions than NDR. 

\vspace{-1.0em}
\subsection{Ablation Study}
\begin{figure}[h]
    \centering
    \includegraphics[trim=00 20 0 00, clip,width=0.75\linewidth]{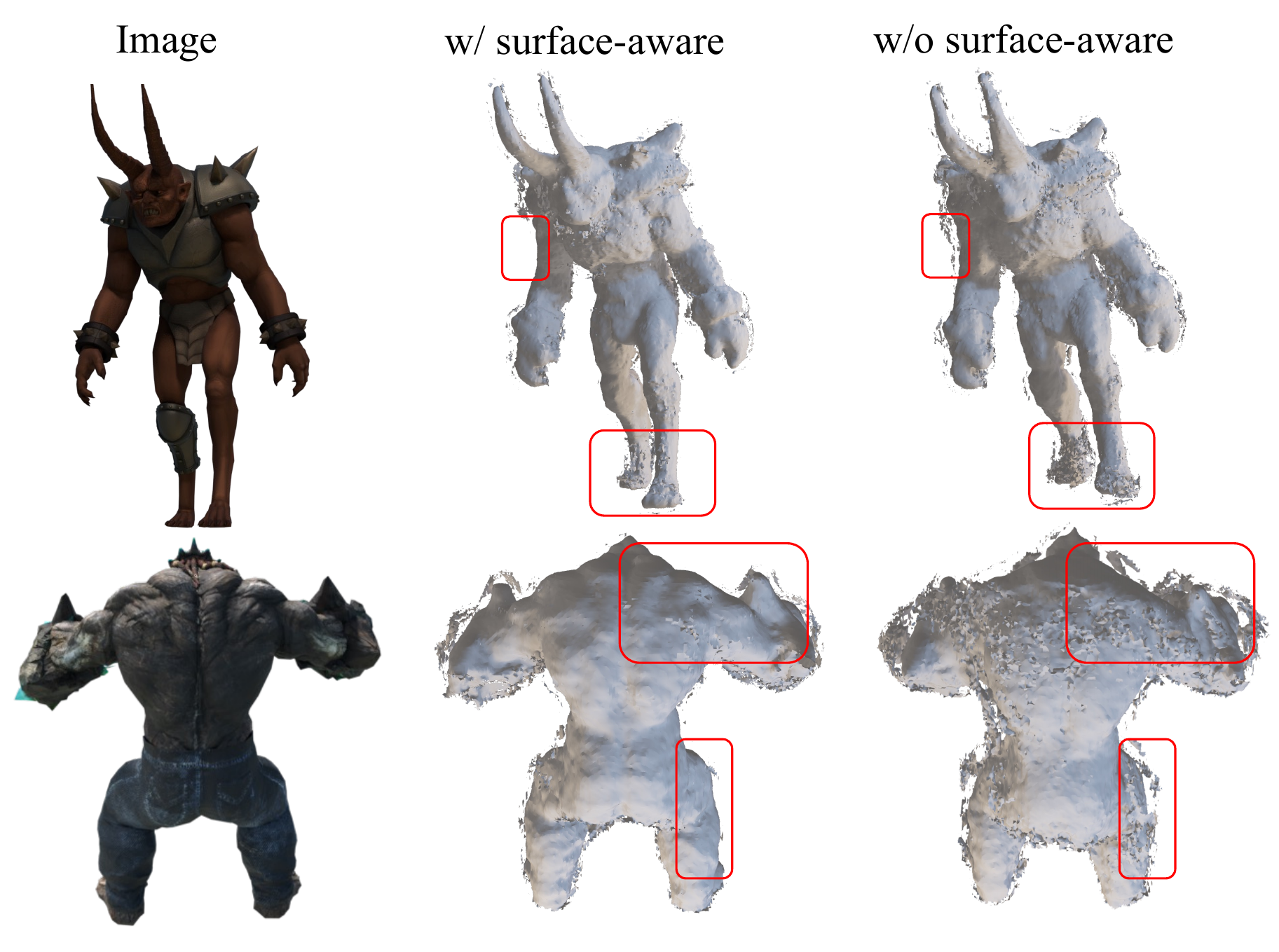}
    \vspace{-1.0em}
    \caption{Demonstration of surface-aware density control. The middle image in each sample shows the mesh from the DGS module with surface-aware density control, while the rightmost image shows the result without density control.}
    \vspace{-1.0em}
    \label{fig:ns_benefits}
\end{figure}

\noindent \textbf{Surface-aware Density Control.}
To illustrate the effect of surface-aware density control, we present two samples of 3D reconstructions with and without surface-aware density control in Fig.~\ref{fig:ns_benefits}. The results show that, in the absence of surface-aware density control, a greater number of points are dispersed away from the surface, resulting in a floating appearance. In contrast, applying surface-aware density control, points are more concentrated and closely aligned with the surface, demonstrating improved reconstruction fidelity and surface adherence.

\vspace{0.5em}
\noindent \textbf{Depth and Normal Supervision.}
The ablation study results in Tab.~\ref{tab:ablation_on_ablation} highlight the complementary effects of depth and normal information on the quality of 3D reconstruction and novel-view synthesis. The depth map includes well-used $\alpha$-blending and filtered depth map as shown in Eq.~(\ref{equ:filtered_depth}). When neither depth nor normal data are used, the model exhibits its lowest performance, underscoring the limitations of relying solely on other cues. Adding depth or normal information individually leads to noticeable improvements, with depth contributing more to spatial alignment accuracy, while normals enhance surface detail and orientation. This suggests that each type of information addresses different aspects of the reconstruction task. As for the effect of the depth map, the filtered depth map showed more promising results than $\alpha$-blending depth maps, highlighting the importance of depth filtering. When both filtered depth and normal data are combined, the model achieves optimal performance across all metrics, indicating the complementary roles of these inputs. Depth data enhances spatial positioning, while normals provide detailed surface cues, resulting in a more accurate and coherent 3D reconstruction. This improved 3D structure also benefits novel-view synthesis by addressing the challenges of monocular dynamic reconstruction, a highly under-constrained optimization problem. Overall, the results emphasize that while depth and normal information offer unique benefits, integrating both is crucial for achieving high fidelity in 3D reconstructions and producing realistic novel views.

\begin{table}[t]
\centering
\caption{Ablation study of depth and normal supervision.}
\vspace{-1em}
\begin{adjustbox}{width=0.40\textwidth}
\begin{tabular}{cccccc}
\hline
 $\alpha$-Depth & Filtered Depth & Normals       & CD $\downarrow$ & EMD $\downarrow$ & PSNR $\uparrow$ \\ \hline
\xmark & \xmark & \xmark & 1.006     & 0.134           &  34.89       \\
\xmark & \xmark & \cmark  & 0.831 & 0.126  & 35.11 \\ 
\cmark & \xmark & \xmark & 0.746 & 0.123  & 34.98 \\  
\xmark & \cmark & \xmark & 0.697 & 0.121  & 35.02 \\  
\cmark & \xmark & \cmark & 0.648 & 0.119  & 35.06 \\  
\xmark &\cmark & \cmark & \textbf{0.502} & \textbf{0.116}  & \textbf{35.14} \\  \hline
\end{tabular}
\end{adjustbox}
\vspace{-1.8em}
\label{tab:ablation_on_ablation}
\end{table}

%% file: sec/5_conclusion.tex
\section{Conclusion}
\label{sec:con}
This paper presented our method, DGNS, a hybrid framework combining Deformable Gaussian Splatting and Dynamic Neural Surfaces to address the challenges of novel-view synthesis and 3D reconstruction in dynamic scenes. Through surface-aware density control, efficient ray-sampling, and depth supervision, our approach leverages interactions between DGS and DNS to achieve state-of-the-art rendering and geometric accuracy.
Experiments on D-NeRF and Dg-mesh datasets demonstrate the robustness of DGNS across complex dynamic scenes, offering a scalable solution for both geometry and appearance modeling. 
\noindent \textbf{Limitations:} This work primarily addresses the concerns around accuracy as in Fig.~\ref{fig:motivation}, although DGNS incorporates efficient ray-sampling, the DNS module still requires substantial computational resources and longer convergence times compared to DGS, making it the primary speed bottleneck in the framework. Additionally, including both modules during the training stage increases the overall memory footprint, which may limit scalability and efficiency, especially on large datasets or scenes with high complexity.
\noindent \textbf{Future work} will focus on enhancing the computational efficiency of MLP-based DNS modules and extending DGNS’s application to handle more complex dynamic scenes.